\documentclass[sigconf,nonacm]{acmart}
\AtBeginDocument{%
  }
\usepackage{amsmath,amsfonts}
\usepackage[table,dvipsnames]{xcolor}
\usepackage{booktabs}
\usepackage{graphicx}
\usepackage{subcaption}
\usepackage{tabularx}
\usepackage{makecell}
\usepackage{multirow}
\usepackage{array}
\usepackage{ragged2e}
\usepackage{enumitem}
\usepackage{balance}
\usepackage{pifont}
\usepackage{calc}
\usepackage{multibib}
\usepackage{pgfplots}
\usepackage{tikz}
\usepackage{algorithmic}
\usepackage[ruled,vlined,linesnumbered]{algorithm2e}
\usepackage{mdframed}
\usetikzlibrary{positioning,fit,backgrounds,shapes.geometric,arrows.meta,calc,decorations.pathreplacing}
\pgfplotsset{compat=1.18}
\graphicspath{{figures/}}
\renewcommand\footnotetextcopyrightpermission[1]{}
\newcommand{\sigSG}{\sigma_{\mathrm{SG}}}        
\newcommand{\sigB}{\sigma_{\mathrm{b}}}          
\newcommand{\sigP}{\sigma_{\mathrm{p}}}          
\newcommand{\gamF}{\gamma_{\mathrm{f}}}          
\newcommand{\gamP}{\gamma_{\mathrm{p}}}          
\newcommand{\ci}[1]{{\scriptsize\color{gray!80!black}[#1]}}  
\newcommand{\dgain}[1]{{\scriptsize\color{ForestGreen}(#1)$^\dagger$}}
\newcommand{\dloss}[1]{{\scriptsize\color{BrickRed}(#1)$^\dagger$}}
\newcommand{\dns}[1]{{\scriptsize\color{gray}(#1)}}

\begin{document}
\raggedbottom

\title{Lesion-Aware Explainable AI for Diabetic Retinopathy Grading from Fundus Images}

\author{Truong Thinh Nguyen}
\authornote{Corresponding author.}
\authornote{Both authors contributed equally to this research.}
\affiliation{%
  \institution{University of Science and Technology of Hanoi}
  \city{Hanoi}
  \country{Vietnam}
}
\email{thinhnt2410944@usth.edu.vn}
\orcid{0009-0006-6248-3783}

\author{Hai Phong Nguyen}
\authornotemark[2]
\affiliation{%
  \institution{University of Science and Technology of Hanoi}
  \city{Hanoi}
  \country{Vietnam}
}
\email{phongnh2410770@usth.edu.vn}

\author{Truong Thanh Hung Nguyen}
\affiliation{%
  \institution{University of New Brunswick}
  \city{Fredericton}
  \state{New Brunswick}
  \country{Canada}
}
\email{hung.ntt@unb.ca}
\orcid{0000-0002-6750-9536}

\author{Giang Son Tran}
\affiliation{%
  \institution{University of Science and Technology of Hanoi}
  \city{Hanoi}
  \country{Vietnam}
}
\email{tran-giang.son@usth.edu.vn}
\orcid{0000-0001-7140-6882}

\author{Hung Cao}
\affiliation{%
  \institution{University of New Brunswick}
  \city{Fredericton}
  \state{New Brunswick}
  \country{Canada}
}
\email{hung.cao@unb.ca}
\orcid{0000-0002-0788-4377}

\renewcommand{\shortauthors}{Nguyen et al.}

\ccsdesc[500]{Applied computing~Life and medical sciences~Health informatics}
\ccsdesc[500]{Computing methodologies~Machine learning~Machine learning approaches~Neural networks}
\ccsdesc[300]{Computing methodologies~Artificial intelligence~Computer vision~Computer vision tasks}

\begin{abstract}
Post-hoc explanations of deep-learning models for diabetic retinopathy (DR) grading are difficult to trust because different attribution methods highlight different retinal regions for the same image, and a visually plausible heatmap does not show that the highlighted pixels drive the prediction. We propose the Consensus-Calibrated Explanation Map (CCEM), which fuses Grad-CAM++, Ada-SISE, and SmoothGrad Integrated Gradients (SG-IG) using image-specific weights derived from annotation-free reliability cues: deletion sensitivity, inter-method agreement, compactness, retinal containment, and peak structure. CCEM explains a frozen ordinal grader, an EfficientNet-B7 with CBAM and a scalar regression head, pre-trained on EyePACS 2015 in Stage~1 and fine-tuned on APTOS 2019 in Stage~2 with quadratic weighted kappa (QWK)-optimized thresholds; it reaches a QWK of 0.920 on the hidden Kaggle test set and 0.559 on the external MAPLES-DR cohort. Explanations are evaluated on MAPLES-DR, 198 MESSIDOR images with expert pixel-level lesion annotations that are never used for training, threshold fitting, or fusion. On the 193 images with annotated lesions, CCEM obtains the highest Soft Pointing Game score (69.61\%; 95\% CI 63.7--75.3), defined as the ratio of the strongest lesion-supported saliency to the global map maximum rather than a lesion detection rate, together with the lowest deletion AUC (0.570) and the best deletion--insertion balance (0.115); each gain is significant against the strongest single explainer on paired tests. Annotation-free, per-image reliability weighting therefore turns three disagreeing explainers into one map whose peak evidence is more lesion-supported and more faithful to the model than that of any single explainer.
\end{abstract}

\keywords{diabetic retinopathy, explainable AI, saliency fusion, ordinal regression, fundus imaging}

\maketitle

\section{Introduction}
\label{sec:introduction}
\begin{figure*}[t]
    \centering
    \input{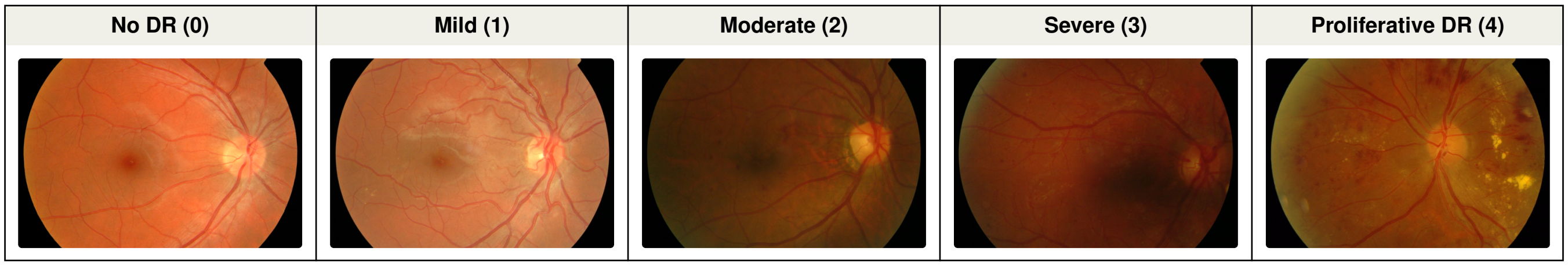}
    \caption{Representative fundus images from the five ordinal DR severity classes.}
    \label{fig:class_examples}
\end{figure*}

Diabetic retinopathy (DR) is a progressive neurovascular complication of diabetes mellitus and a major cause of preventable visual impairment among working-age adults~\cite{teo2021}. Approximately 23\% of people with diabetes develop DR, and around 11\% develop sight-threatening disease requiring timely intervention~\cite{owens2025}. Because early DR is often asymptomatic, regular retinal screening is essential for detecting pathological changes before irreversible vision loss occurs.
Color fundus photography is widely used for DR screening because it is non-invasive, relatively inexpensive, and can capture characteristic retinal findings such as microaneurysms, hemorrhages, exudates, cotton-wool spots, vascular abnormalities, and neovascularization~\cite{wilkinson2003}. Under the International Clinical Diabetic Retinopathy (ICDR) severity scale, fundus images are assigned to five ordered grades: No DR, Mild, Moderate, Severe, and Proliferative DR~\cite{wilkinson2003} (see Figure~\ref{fig:class_examples}). Manual grading, however, is time-consuming, requires specialist expertise, and is subject to inter-observer variability, limiting the scalability of screening programs in resource-constrained settings~\cite{gulshan2016}.

Deep learning (DL) has substantially advanced automated DR grading by learning discriminative retinal representations directly from fundus images~\cite{gulshan2016}. Because DR grades are ordinal, regression-based formulations are well suited to the task: they model disease severity continuously and distinguish adjacent-grade errors from more severe misclassifications.
Despite strong benchmark performance, two challenges remain. First, model accuracy may decline across datasets because of differences in cameras, illumination, image processing, patient populations, grade distributions, and annotation protocols. Second, deep neural networks remain difficult to interpret, creating uncertainty about whether predictions are based on clinically meaningful retinal evidence or on irrelevant image characteristics.

Post-hoc explainable artificial intelligence (XAI) methods address the second challenge by highlighting regions associated with a prediction. However, different attribution mechanisms may produce markedly different heatmaps for the same image \cite{krishna2022disagreement}. Activation-based methods often yield broad but coarse localization \cite{chattopadhay2018grad}, perturbation-based methods emphasize compact regions that influence the output \cite{sudhakar2021ada}, and input-gradient methods provide finer but sometimes diffuse attributions \cite{smilkov2017smoothgrad}. A single explainer, therefore, offers only a partial view of the model's decision process, and presenting several explainers side by side shifts the task of reconciling them onto the reader. Explanation quality should also be assessed across complementary dimensions: lesion overlap reflects clinical localization, deletion and insertion analyses assess model faithfulness, and validity diagnostics identify empty maps or undefined perturbation curves.

This paper addresses the reconciliation problem with the Consensus-Calibrated Explanation Map (CCEM). CCEM fuses Grad-CAM++ \cite{chattopadhay2018grad}, Ada-SISE \cite{sudhakar2021ada}, and SmoothGrad Integrated Gradients (SG-IG) \cite{smilkov2017smoothgrad}, representing activation-, perturbation-, and input-gradient-based attribution, into a single map whose weights are recomputed for every image from annotation-free reliability cues. The method is evaluated on a frozen ordinal DR grader, an EfficientNet-B7 with a Convolutional Block Attention Module (CBAM) and a scalar regression head, trained in two stages (EyePACS 2015 pre-training, APTOS 2019 fine-tuning) with quadratic-weighted kappa (QWK)-optimized decision thresholds. Explanations are evaluated on MAPLES-DR~\cite{lepetit2024}, a public dataset of 198 MESSIDOR fundus images with expert pixel-level annotations of DR lesions and image-level grades, which is reserved exclusively for external inference and post-hoc evaluation.
The main contributions of our work are:

\begin{enumerate}
    \item \textbf{We propose CCEM}, an adaptive explanation-fusion method that assigns image-specific reliability weights to Grad-CAM++, Ada-SISE, and SG-IG using five annotation-free cues (deletion sensitivity, agreement, compactness, retinal containment, and peak structure) and a temperature-controlled softmax, so that no expert mask enters the fusion.

    \item \textbf{We establish an explanation-evaluation protocol} that jointly reports expert-lesion localization, perturbation-based model faithfulness, and numerical validity on the same external cohort, with per-image paired statistics.

    \item \textbf{We provide the ordinal grading substrate for this study}, a two-stage EfficientNet-B7+CBAM regression model with QWK-optimized thresholds, reported with internal, hidden-test, and external cross-dataset performance so that explanation quality can be read against the grader's own domain shift.
\end{enumerate}

\section{Related Work}
\label{sec:related_work}

\begin{table*}[t]
\centering
\caption{Methodological comparison of representative explainable DR frameworks. ``Train'' and ``Eval.'' indicate whether expert lesion annotations are used to supervise the model and to score the explanations, respectively. \ding{55} indicates that the corresponding evaluation dimension was not reported in the cited study.}
\label{tab:dr-xai-comparison}
\small
\setlength{\tabcolsep}{3.5pt}
\renewcommand{\arraystretch}{1.12}
\begin{tabularx}{\textwidth}{>{\raggedright\arraybackslash}p{2.75cm}
  >{\raggedright\arraybackslash}X
  cc
  >{\raggedright\arraybackslash}p{1.95cm}
  >{\raggedright\arraybackslash}p{1.75cm}
  >{\raggedright\arraybackslash}p{3.0cm}}
\toprule
& & \multicolumn{2}{c}{\textbf{Lesion masks}} & \multicolumn{3}{c}{\textbf{Explanation evaluation}} \\
\cmidrule(lr){3-4} \cmidrule(l){5-7}
\textbf{Study} & \textbf{Explanation design} & \textbf{Train} & \textbf{Eval.} & \textbf{Localization} & \textbf{Faithfulness} & \textbf{Validity/coverage} \\
\midrule
ExplAIn~\cite{quellec2021} & Intrinsic lesion representations learned jointly with image-level grading & No & Yes & Pixel-level & \ding{55} & \ding{55} \\
\rowcolor{gray!7} DR\textbar GRADUATE~\cite{araujo2020} & Joint ordinal grading, attention explanation, and uncertainty & No & No & Qualitative & \ding{55} & \ding{55} \\
Chetoui and Akhloufi~\cite{chetoui2020} & Model-specific spatial explanations for end-to-end DR prediction & No & No & Qualitative & \ding{55} & \ding{55} \\
\rowcolor{gray!7} Herrero-Tudela et~al.~\cite{herrero2025} & SHAP analysis of a modified ResNet-50 grader across datasets & No & No & Visual/clinical & \ding{55} & \ding{55} \\
ResViT FusionNet~\cite{ikram2025} & LIME and Grad-CAM for a CNN\nobreakdash--Transformer grader & No & Yes & Qualitative & \ding{55} & \ding{55} \\ 
\midrule
\textbf{CCEM (Ours)} & Grad-CAM++, Ada-SISE, and SG-IG with image-specific reliability-calibrated fusion & No & Yes & EBPG, SoftPG & Del., ins., OA & Valid curves, undefined cases, zero maps \\
\bottomrule
\end{tabularx}
\end{table*}

Automated DR grading has progressed from conventional convolutional classifiers toward attention-enhanced, ordinal, uncertainty-aware, and hybrid architectures. EfficientNet-based models use compound scaling to balance network depth, width, and input resolution~\cite{tan2019efficientnet}, attention modules such as CBAM recalibrate feature responses~\cite{woo2018cbam}, and CNN--Transformer architectures combine local lesion patterns with broader spatial context~\cite{ikram2025}. DR|GRADUATE jointly estimates an ordinal grade, an attention-based explanation, and predictive uncertainty to support specialist review~\cite{araujo2020}. Herrero-Tudela et al. evaluated a modified ResNet-50 grader across several retinal datasets with SHAP~\cite{herrero2025}, whereas ResViT FusionNet combined ResNet-50 and Vision Transformer representations with LIME and Grad-CAM explanations~\cite{ikram2025}. These studies integrate transfer learning, attention, uncertainty, and interpretability, but their primary objective is not the adaptive fusion of multiple post-hoc attribution families.

Existing DR-XAI approaches include intrinsic lesion representations, class-activation mapping, perturbation-based explanations, and feature attribution. Chetoui and Akhloufi visualized retinal regions associated with DR predictions across multiple datasets~\cite{chetoui2020}. ExplAIn derives image-level predictions from intermediate lesion representations learned under image-level supervision and evaluates them at both image and pixel levels~\cite{quellec2021}. Other studies have applied SHAP, LIME, and Grad-CAM to examine the retinal evidence supporting model decisions~\cite{herrero2025,ikram2025}. However, these explanations are generally assessed individually or presented in parallel rather than reconciled. Evaluation practices are also inconsistent: expert-lesion localization and perturbation-based faithfulness are seldom examined together, and empty maps or undefined perturbation measurements are rarely reported. Table~\ref{tab:dr-xai-comparison} compares representative DR-XAI frameworks by explanation design, use of expert masks for training and for evaluation, faithfulness evaluation, and validity reporting; predictive accuracy is omitted because the studies use different datasets, grading schemes, and validation protocols.

Outside ophthalmology, the instability of individual attribution methods has motivated explicit aggregation. Rieger and Hansen showed that averaging several explanation methods yields maps that are more stable under input perturbation and more robust to adversarial manipulation than any constituent method~\cite{rieger2019aggregating}. Krishna et al. documented that popular explainers frequently disagree on the same prediction and that practitioners lack a principled rule for choosing among them~\cite{krishna2022disagreement}. Mahlau and Nolde optimized fixed aggregation weights for saliency maps against faithfulness criteria~\cite{mahlau2022fidelity}. Closest to our work, Somani et al. weighted several saliency maps per image by their concentration, inter-explainer agreement, and the quality of the resulting SAM2 segmentation, and used the fused maps as pseudo-labels for weakly supervised segmentation~\cite{somani2025explain}. CCEM shares the per-image weighting principle but differs in purpose and in its reliability signal: it explains a clinical grader rather than producing segmentation labels, includes a deletion-sensitivity factor that queries the explained model directly, and is evaluated jointly against expert lesion annotations and perturbation-based faithfulness on an external DR cohort.

\section{Data and Preprocessing}

\subsection{Datasets and Data Split}
\label{sec:datasets}
This study draws on three publicly available color fundus photography datasets, i.e., EyePACS 2015~\cite{eyepacs2015diabetic}, APTOS 2019~\cite{aptos2019blindness}, and MAPLES-DR~\cite{lepetit2024}, each assigned a distinct, non-overlapping role. EyePACS 2015 served as the source domain for Stage~1 pre-training, APTOS 2019 as the target domain for Stage~2 fine-tuning and internal validation, and MAPLES-DR, which uniquely provides pixel-level expert lesion annotations, was reserved exclusively for external cross-dataset evaluation and explainability analysis. Because the three datasets originate from different acquisition sites, camera systems, and image distributions, this design also enables explicit examination of domain shift between internal and external evaluation conditions. All splits were drawn at the image level: EyePACS file names encode a patient identifier and laterality, so the two eyes of one patient may fall on different sides of the Stage~1 split, whereas APTOS 2019 does not release patient identifiers, so a patient-level partition is not possible for that dataset. No deduplication was applied to either training corpus. 

\textbf{EyePACS 2015} contains 35,126 fundus images labeled across the five ordinal DR severity levels~\cite{eyepacs2015diabetic} and is markedly imbalanced, with No DR accounting for the majority of images (Table~\ref{tab:dataset_distribution}). It was partitioned by stratified sampling into 90\% for pre-training and 10\% for validation, preserving the original class proportions. The EyePACS validation split was used only to select the Stage~1 checkpoint; its thresholds were not transferred, and no result reported in this paper is computed on EyePACS.

\begin{table}[t]
\centering
\caption{Distribution of retinal images across DR severity levels in EyePACS 2015 and APTOS 2019.}
\label{tab:dataset_distribution}
\footnotesize
\setlength{\tabcolsep}{0pt}
\begin{tabular*}{\columnwidth}{@{\extracolsep{\fill}}lcccccc@{}}
\toprule
& \multicolumn{3}{c}{\textbf{EyePACS 2015}} & \multicolumn{3}{c}{\textbf{APTOS 2019}} \\
\cmidrule(lr){2-4} \cmidrule(l){5-7}
\textbf{Class} & \textbf{Total} & \textbf{Pre-train} & \textbf{Val.} & \textbf{Total} & \textbf{Fine-tune} & \textbf{Val.} \\
\midrule
No DR & 25,810 & 23,229 & 2,581 & 1,805 & 1,444 & 361 \\
Mild & 2,443 & 2,199 & 244 & 370 & 296 & 74 \\
Moderate & 5,292 & 4,763 & 529 & 999 & 799 & 200 \\
Severe & 873 & 786 & 87 & 193 & 154 & 39 \\
Proliferative DR & 708 & 636 & 72 & 295 & 236 & 59 \\
\midrule
\textbf{Total} & \textbf{35,126} & \textbf{31,613} & \textbf{3,513} & \textbf{3,662} & \textbf{2,929} & \textbf{733} \\
\bottomrule
\end{tabular*}
\end{table}

\begin{table}[t]
\centering
\caption{MAPLES-DR grades, ICDR mapping, and image counts for external grading (198) and XAI evaluation (193 with a non-empty expert lesion mask).}
\label{tab:maples_distribution}
\footnotesize
\setlength{\tabcolsep}{0pt}
\begin{tabular*}{\columnwidth}{@{\extracolsep{\fill}}llcc@{}}
\toprule
\textbf{MAPLES-DR grade} & \textbf{ICDR grade} & \textbf{Grading} & \textbf{XAI} \\
\midrule
R0 (no retinopathy) & 0 No DR & 14 & 9 \\
R1 (mild) & 1 Mild & 155 & 155 \\
R2 (moderate) & 2 Moderate & 15 & 15 \\
R3 (severe) & 3 Severe & 10 & 10 \\
R4A (proliferative, active) & 4 Proliferative DR & 4 & 4 \\
R4S (proliferative, stable) & 4 Proliferative DR & 0 & 0 \\
R6 (ungradable) & excluded & 0 & 0 \\
\midrule
\textbf{Total} & & \textbf{198} & \textbf{193} \\
\bottomrule
\end{tabular*}
\end{table}

\textbf{APTOS 2019}\label{sec:aptos} comprises 3,662 fundus images labeled across five ordinal DR grades~\cite{aptos2019blindness}. The images were partitioned by image identifier into five stratified folds; fold~0 (733 images, 20\%) served as the validation subset and the remaining four folds (2,929 images, 80\%) as the fine-tuning subset. The validation subset was used for epoch-level model selection, Nelder--Mead threshold optimization, and internal performance reporting. Consequently, the APTOS results reported in this study are internal validation results rather than performance on a fully independent test set; the APTOS 2019 Kaggle private test set provides the held-out estimate (Section~\ref{sec:internal_results}).

\textbf{MAPLES-DR}\label{sec:maples} was selected for external evaluation because, unlike EyePACS 2015 and APTOS 2019, it provides expert pixel-wise annotations of retinal biomarkers in addition to image-level diagnostic information. The dataset comprises 198 fundus images (138 in its official training split and 60 in its test split) drawn from the MESSIDOR collection~\cite{lepetit2024}; both official splits were pooled because no MAPLES-DR image is used for training in this study. DR grades are supplied on a six-level scale derived from the Canadian teleophthalmology screening protocol (R0--R3, R4A, R4S, and R6 for ungradable images); Table~\ref{tab:maples_distribution} gives the mapping to the five ICDR grades used by the grader and the resulting distribution. Only R0--R4A occur in the dataset, and Mild DR (R1) accounts for 155 of the 198 images. Five images, all graded R0, have no annotated lesion and therefore an empty consolidated mask (Section~\ref{sec:mask_construction}); they are retained for external grading (198 images) and excluded from the mask-based explanation metrics (193 images). The image-level labels were used as external reference labels for cross-dataset grading context and were not treated as incontrovertible clinical ground truth. No MAPLES-DR image-level labels or expert lesion masks were used for classifier training, CBAM optimization, decision-threshold fitting, or tuning the explanation pipeline. MAPLES-DR images were used exclusively as external inference inputs for generating explanations and estimating image-specific CCEM reliability weights, while expert lesion masks were reserved for post-hoc localization evaluation.

\subsection{Data Preprocessing}
\label{sec:preproc}
All fundus images were passed through a standardized preprocessing pipeline to suppress sources of non-pathological variation, such as uneven illumination, black borders, and inconsistent framing across acquisition devices, while preserving the subtle chromatic and textural cues on which DR grading depends. Each image was first converted to grayscale and thresholded at \(\texttt{tol}=7\) to identify the illuminated retinal region, and the smallest bounding box containing all foreground pixels was applied to the three color channels; if no valid foreground was detected, the original image was retained. Contrast-Limited Adaptive Histogram Equalization (CLAHE)~\cite{zuiderveld1994contrast} was then applied to the luminance channel in CIE-LAB space with an \(8 \times 8\) tile grid and a contrast limit of 2.0, leaving the chromatic channels unchanged. Finally, each image was resized to $600 \times 600$ pixels, the native input resolution of EfficientNet-B7, and normalized with the ImageNet channel means $(0.485, 0.456, 0.406)$ and standard deviations $(0.229, 0.224, 0.225)$, matching the domain on which the pretrained filters were learned. During training, these deterministic steps were followed by the stochastic augmentation policy of Section~\ref{sec:training}; validation and inference used deterministic preprocessing, except for the test-time augmentation described in Section~\ref{sec:training}.

\subsubsection{MAPLES-DR Expert Lesion Mask Construction}
\label{sec:mask_construction}
For the localization-based explainability evaluation, the expert segmentations of microaneurysms, hemorrhages, exudates, and cotton-wool spots supplied by MAPLES-DR were combined into one consolidated expert lesion mask by pixel-wise logical union,
\begin{equation}
    M_{\text{expert}} = M_{\text{MA}} \vee M_{\text{HE}} \vee M_{\text{EX}} \vee M_{\text{CW}},
    \label{eq:mask_union}
\end{equation}
such that a pixel is labeled lesion-positive if it is annotated in any of the four constituent masks. Each mask was resampled with nearest-neighbor interpolation, preserving its binary structure, and cropped with the same spatial transformation as the corresponding fundus image, so that expert references and explanation maps share a common coordinate system. The consolidated masks were used only to compute the localization metrics of Section~\ref{sec:evaluation} and did not enter the classifier objective, explainer construction, post-processing pipeline, or CCEM reliability score.

\section{Proposed Framework}
This section describes the five-grade DR grader that serves as the substrate for the explanation study (Sections~\ref{sec:network}--\ref{sec:training}), the proposed CCEM (Section~\ref{sec:ccem}), and the evaluation protocol (Section~\ref{sec:evaluation}). MAPLES-DR is used only for external inference and explanation evaluation, not for training, threshold optimization, or CCEM weight estimation. Table~\ref{tab:notation} lists every tunable constant of the explanation pipeline with the symbol used for it in the text.

\begin{figure*}[t]
    \centering
    \input{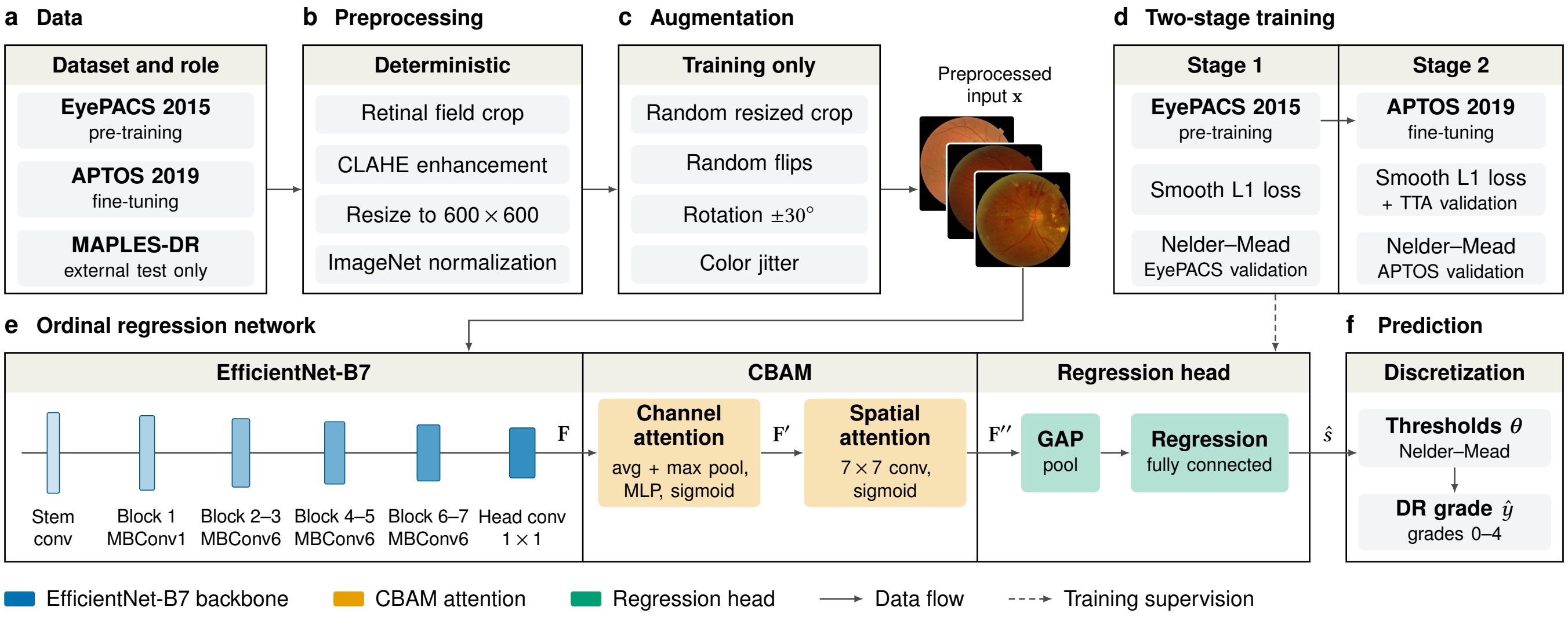}
    \caption{End-to-end architecture of the EfficientNet-B7+CBAM ordinal grading model. EyePACS 2015 and APTOS 2019 enter the two-stage training path; MAPLES-DR is passed through the frozen model for inference only and never supplies a training signal. The convolutional backbone and attention module produce a continuous severity score that is converted to one of five ordered DR grades by validation-optimized thresholds.}
    \label{fig:complete_architecture}
\end{figure*}

\begin{table}[t]
\centering
\caption{Notation and fixed constants of the explanation pipeline. Each symbol has a single meaning throughout the paper. All values were set by hand before the MAPLES-DR evaluation and were not tuned on MAPLES-DR images or masks.}
\label{tab:notation}
\small
\setlength{\tabcolsep}{4pt}
\begin{tabular}{llc}
\toprule
\textbf{Symbol} & \textbf{Meaning} & \textbf{Value} \\
\midrule
$\sigSG$ & SmoothGrad noise std (normalized input space) & 0.10 \\
$N_{\mathrm{SG}}$ & SmoothGrad samples (squared variant) & 16 \\ 
$K_{\mathrm{IG}}$ & Integrated Gradients path steps & 24 \\ 
\midrule
$q$ & keep percentile in per-map sparsification & 94 \\ 
$\gamP$ & per-map sharpening exponent & 2.5 \\ 
$\sigP$ & per-map Gaussian blur (px) & 0.6 \\ 
$\lambda$ & attribution fraction retained under the lesion prior & 0.20 \\ 
\midrule
$\rho$ & pixel fraction removed in deletion sensitivity $F_m$ & 5\% \\
$\tau$ & softmax temperature over reliability scores & 0.25 \\
$\sigB$ & Gaussian blur of the fused map (px) & 1.0 \\
$\gamF$ & sharpening exponent of the fused map & 1.5 \\
$(p_{\mathrm{lo}}, p_{\mathrm{hi}})$ & percentiles of robust normalization & (1, 99) \\
\midrule
$\epsilon$ & numerical guard & $10^{-8}$ \\
\bottomrule
\end{tabular}
\end{table}

\subsection{Network Architecture}
\label{sec:network}
The grader combines a high-capacity convolutional backbone, a lightweight attention module, and a single-output ordinal regression head (Figure~\ref{fig:complete_architecture}). The backbone was instantiated with the \texttt{timm} \texttt{tf\_efficientnet\_b7\_ns} architecture, initialized with Noisy Student weights~\cite{xie2020self}, and stripped of its classification layer. Compound scaling gives EfficientNet-B7 a native input resolution of $600 \times 600$ pixels and a deep stack of MBConv blocks with squeeze-and-excitation units~\cite{tan2019efficientnet}, making it well suited to resolving the fine-grained, low-contrast lesions that separate adjacent DR grades. For an input $x \in \mathbb{R}^{3 \times 600 \times 600}$, the backbone produces a final feature map $\mathbf{F} \in \mathbb{R}^{C \times H \times W}$ with $C=2560$.

A CBAM~\cite{woo2018cbam} is applied once to $\mathbf{F}$, immediately before global average pooling (GAP). It first rescales channels with a sigmoid-gated attention vector computed from average- and max-pooled channel descriptors through a shared bottleneck multilayer perceptron (reduction ratio $r=16$), and then rescales spatial positions with a sigmoid-gated map obtained by a $7 \times 7$ convolution over channel-wise average and max pooling, yielding the attention-refined map $\mathbf{F}''$. CBAM requires no lesion-level supervision and should therefore be interpreted as learned feature recalibration rather than an explicit lesion-localization mechanism.

\subsubsection{Ordinal Regression Head and Threshold Calibration}
\label{sec:ordinal_head}
\label{sec:thresholds}
DR severity labels possess a natural ordering, and errors between distant grades are more consequential than errors between adjacent grades. The network was therefore formulated as a single-output regression model: $\mathbf{z} = \text{GAP}(\mathbf{F}'') \in \mathbb{R}^C$ is mapped to a scalar severity score $\hat{s} = \mathbf{w}^\top \mathbf{z} + b$ and trained against the reference grade $s$ with the Smooth L1 (Huber) objective,
\begin{equation}
\mathcal{L}_{\text{Smooth L1}}(\hat{s}, s) =
    \begin{cases}
        \frac{1}{2}(\hat{s} - s)^2, & |\hat{s} - s| < 1, \\
        |\hat{s} - s| - \frac{1}{2}, & \text{otherwise},
    \end{cases}
    \label{eq:smooth_l1}
\end{equation}
which combines quadratic sensitivity to small residuals with reduced sensitivity to large ones. The continuous score is converted to an ordinal grade with four ordered decision thresholds $\boldsymbol{\theta} = (\theta_1, \ldots, \theta_4)$ rather than naive rounding,
\begin{equation}
    \hat{y} = \textstyle\sum_{k=1}^{4} \mathbf{1}[\hat{s} \ge \theta_k], \qquad
    \boldsymbol{\theta}^\star = \arg\max_{\boldsymbol{\theta}} \text{QWK}(y, \hat{y}(\hat{s}; \boldsymbol{\theta})),
    \label{eq:thresholds}
\end{equation}
where the thresholds are initialized at the midpoints (0.5, 1.5, 2.5, 3.5) and optimized on the validation predictions with the Nelder--Mead simplex algorithm~\cite{nelder1965simplex} to maximize QWK~\cite{cohen1968weighted}. Nelder--Mead was selected because hard thresholding makes QWK non-differentiable with respect to $\boldsymbol{\theta}$. This separates continuous severity representation learning from the final discretization rule. Thresholds were fitted independently in each training stage; the EyePACS thresholds were not transferred, and the final APTOS thresholds were saved with the selected checkpoint for use at test time.

\subsection{Model Training}
\label{sec:training}
\label{sec2.4}
\label{sec:tta}
The model was trained in two sequential stages to separate large-scale source-domain representation learning from target-domain adaptation. Both stages used the preprocessing of Section~\ref{sec:preproc}, the objective of Eq.~\eqref{eq:smooth_l1}, and AdamW~\cite{loshchilov2019decoupled} with a weight decay of $1 \times 10^{-5}$. The staged procedure is intended to reduce dependence on the limited target-domain sample size; it does not, by itself, eliminate overfitting or domain shift.

\paragraph{Stage 1: source-domain pre-training on EyePACS 2015.} The backbone, CBAM, and regression head were trained end-to-end on the EyePACS pre-training split for ten epochs with a learning rate of $1 \times 10^{-4}$ and a batch size of 32. At the end of each epoch, validation predictions were discretized with the threshold search of Section~\ref{sec:thresholds}, and the checkpoint with the highest validation QWK was retained. Only its weights were transferred to Stage~2.

\paragraph{Stage 2: target-domain fine-tuning on APTOS 2019.} Training resumed from the Stage~1 weights on the APTOS fine-tuning subset for 20 epochs with a batch size of 16. The learning rate was lowered to $2 \times 10^{-5}$ to make more conservative updates to the pre-trained representation and followed a cosine annealing schedule to a floor of $1 \times 10^{-6}$; automatic mixed-precision computation reduced the memory footprint at the $600 \times 600$ resolution. At each epoch, the threshold vector was optimized on the same TTA-averaged validation scores used to compute QWK, and the checkpoint and threshold vector that jointly achieved the highest validation QWK were retained.

\paragraph{Class imbalance and augmentation.} The final pipeline did not use class-weighted loss, focal loss, weighted sampling, or explicit oversampling. Imbalance was addressed indirectly through stratified splitting, ordinal regression, validation-based threshold calibration, and stochastic augmentation consisting of random resized cropping (scale 0.8--1.0), horizontal flipping, vertical flipping where enabled by the stage implementation, rotation up to $\pm 30^\circ$, and random brightness, contrast, saturation, and hue perturbations. These measures do not remove the underlying grade imbalance, and residual minority-grade errors are reported explicitly in the class-wise results.

\paragraph{Test-time augmentation (TTA)} Inference averages the continuous outputs of three deterministic views, the original image and its horizontal and vertical flips, $\hat{s}_{\text{TTA}} = \frac{1}{3}(\hat{s}_{\text{orig}} + \hat{s}_{\text{hflip}} + \hat{s}_{\text{vflip}})$, before thresholding. The same protocol was used for Stage~2 checkpoint selection, threshold fitting, and internal reporting. External grading on MAPLES-DR and all explanation experiments used the single deterministic view, so that every explainer and every perturbation test refer to the same input tensor.

\subsection{Consensus-Calibrated Explanation Map}
\label{sec:ccem}
\subsubsection{Multi-XAI Explanation Generation}
The final Stage~2 model was frozen before explanation generation so that every method described the same learned predictor. We selected three attribution families with different mechanisms: Grad-CAM++~\cite{chattopadhay2018grad}, Ada-SISE~\cite{sudhakar2021ada}, and SG-IG~\cite{sundararajan2017axiomatic,smilkov2017smoothgrad}. All explainers target the scalar severity output before thresholding, which avoids explanation discontinuities introduced by the ordinal decision thresholds and keeps the attribution target consistent across the five predicted grades.

Grad-CAM++ operates on the final attention-refined convolutional feature map and combines spatial activations with gradient-derived channel weights; relative to standard Grad-CAM, its pixel-wise weighting is intended to represent multiple discriminative regions more effectively. Ada-SISE forms spatial masks from informative activation channels at multiple EfficientNet stages and scores each mask through its effect on the model output, providing a perturbation-oriented view. SG-IG applies Integrated Gradients along a path of $K_{\mathrm{IG}}$ steps from a zero baseline in normalized input space and stabilizes the resulting pixel attributions with SmoothGrad-squared averaging over $N_{\mathrm{SG}}$ Gaussian-perturbed copies of the input with noise standard deviation $\sigSG=0.10$.

The raw outputs of the three methods differ in spatial resolution, sparsity, and dynamic range. A shared compact post-processing stage therefore converts every explanation into a non-negative map $H_m\in[0,1]^{h\times w}$, resizes it to the common $600\times600$ evaluation grid, and masks pixels outside the retinal field of view. Each map is min-max normalized, sparsified by retaining values above its $q$-th percentile, sharpened with exponent $\gamP$, and lightly smoothed with a Gaussian of standard deviation $\sigP$ to suppress low-amplitude background responses without destroying local structure. Two weak image-derived corrections are also used: optic-disc suppression reduces a common bright-structure confound, and a lesion-candidate appearance prior mildly favors local chromatic/textural irregularities by multiplying the map with $\lambda+(1-\lambda)\,\pi(x)$, where $\pi(x)\in[0,1]$ is the prior and $\lambda=0.20$ guarantees that a fixed fraction of the original attribution is always retained. Both operations are annotation-free and are applied identically to the three baselines and to the CCEM inputs, so that they do not favor the fused map. If any operation creates an empty or non-finite map, the pipeline falls back to the most recent valid representation.

\begin{figure*}[t]
    \centering
    \input{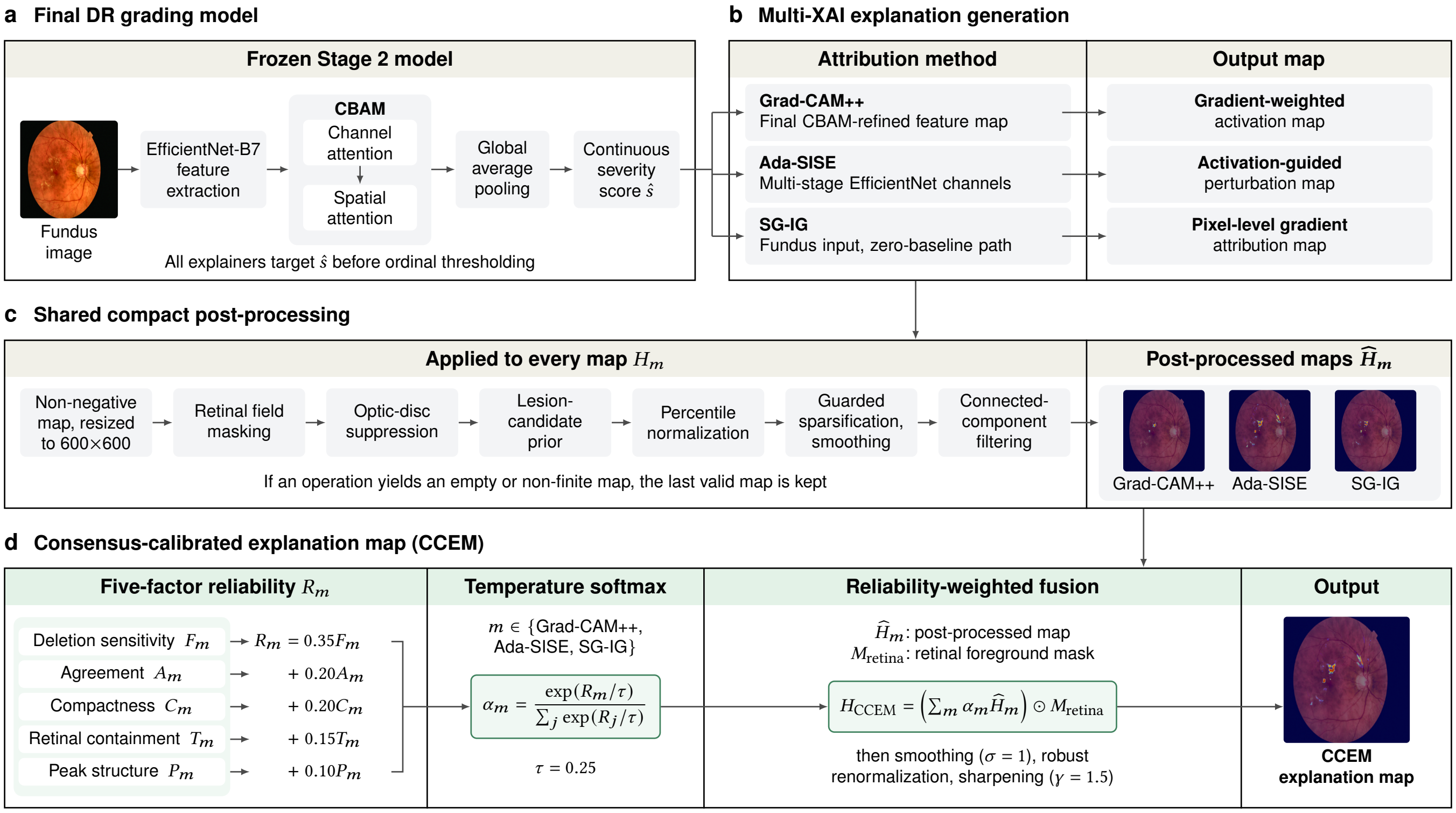}
    \caption{CCEM Explanation Method. The frozen grader produces three complementary attribution maps, which undergo shared compact post-processing, image-specific reliability estimation, adaptive fusion, and external evaluation with expert masks and perturbation tests.}
    \label{fig:xai_pipeline}
\end{figure*}

\subsubsection{Image-Specific Reliability Estimation}
A fixed average assumes that every explainer is equally informative on every image. In practice, an attribution method can be compact and influential on one image but diffuse, unstable, or dominated by non-retinal structure on another. CCEM therefore estimates reliability separately for each explanation map and each input image, using only quantities available from the image, the model, and the explanation maps. Before scoring, each post-processed map $\widehat H_m$ is brought to a common scale by robust percentile normalization: values are clipped to the 1st and 99th percentiles of its positive pixels and rescaled to $[0,1]$, which prevents isolated extreme values from dominating the comparison between methods.

For explainer $m$, the reliability score is
\begin{equation}
R_m=0.35F_m+0.20A_m+0.20C_m+0.15T_m+0.10P_m,
\label{eq:ccem_reliability}
\end{equation}
where every factor lies in $[0,1]$ and the coefficients sum to one. The coefficients were set by hand, with the largest weight on the only factor that queries the model, and were fixed before the MAPLES-DR evaluation rather than fitted to its lesion labels.

\textbf{Deletion sensitivity} $F_m$ estimates whether the most salient pixels affect the model output. The top $\rho=5\%$ of pixels ranked by $\widehat H_m$ are replaced by the normalized zero baseline, and the relative change in the continuous severity score is converted to a bounded score; larger output degradation gives larger reliability. \textbf{Agreement} $A_m$ is the mean pairwise Spearman rank correlation between $\widehat H_m$ and the other two maps after spatial alignment, rescaled to $[0,1]$. Agreement is deliberately only one component, because consensus among explainers does not imply clinical correctness. \textbf{Compactness} $C_m$ uses Hoyer sparsity~\cite{hoyer2004non} so that maps with concentrated evidence are preferred over nearly uniform responses while still allowing multiple lesion regions. \textbf{Retinal containment} $T_m$ is the fraction of saliency mass inside the detected retinal field. \textbf{Peak structure} $P_m$ compares high-percentile saliency with the map mean and rewards explanations with a clear but not necessarily single dominant response.

\subsubsection{Adaptive Fusion}
The reliability scores are converted into image-specific weights with a temperature-controlled softmax,
\begin{equation}
\alpha_m=\frac{\exp(R_m/\tau)}{\sum_j\exp(R_j/\tau)},\qquad \tau=0.25,
\label{eq:ccem_weights}
\end{equation}
which preserves contributions from all valid explainers while emphasizing more reliable maps: with $\tau=0.25$, a reliability gap of 0.1 between two maps changes their weight ratio by a factor of $e^{0.4}\approx1.5$, whereas $\tau\to\infty$ would recover equal weighting and $\tau\to0$ would select a single map. The fused map is
\begin{equation}
H_{\mathrm{CCEM}}=\left(\sum_m\alpha_m\widehat H_m\right)\odot M_{\mathrm{retina}},
\label{eq:ccem_fusion}
\end{equation}
where $M_{\mathrm{retina}}$ is the retinal foreground mask. The fused result is smoothed with a Gaussian of standard deviation $\sigB=1.0$ pixel, robustly renormalized, sharpened with exponent $\gamF=1.5$, and robustly renormalized once more; the blur removes the resampling seams between maps of different native resolution, and the exponent restores the contrast that averaging flattens. If the final map is empty or invalid, CCEM returns the highest-reliability non-empty source map. These safeguards guarantee a usable visualization when at least one source explainer succeeds, but they do not alter the classifier output.

The design should be interpreted as reliability-calibrated evidence aggregation rather than a lesion detector. In particular, CCEM cannot introduce pathology that is absent from all source explanations, and agreement can reinforce shared model bias. The external evaluation therefore separates expert-lesion localization from perturbation-based faithfulness and reports both alongside the grader's cross-dataset performance.

\subsection{Evaluation Protocol}
\label{sec:evaluation}
Evaluation was divided into predictive grading, explanation localization, explanation faithfulness, and numerical validity. Internal grading metrics were computed on the 733-image APTOS validation fold; because this fold was also used for checkpoint selection and threshold optimization, they are reported as internal validation rather than as an independent test estimate. External grading context was obtained by applying the frozen Stage~2 model and the APTOS-derived thresholds to all 198 MAPLES-DR images without recalibration. We report exact-grade accuracy and QWK, with macro and weighted classification summaries for the internal set. The Kaggle private-test QWK is reported separately because its hidden labels do not permit reconstruction of accuracy or class-wise statistics.

\paragraph{Localization metrics.} For XAI evaluation, the consolidated expert mask of Section~\ref{sec:mask_construction} defines the 193-image cohort shared by every explanation method. Let $H\in[0,1]^{h\times w}$ be a non-negative explanation rescaled by its maximum and $M\in\{0,1\}^{h\times w}$ the expert lesion mask. Energy-Based Pointing Game (EBPG)~\cite{wang2020score} measures the fraction of total saliency energy falling inside the expert mask,
\begin{equation}
\mathrm{EBPG}=\frac{\sum_p H_pM_p}{\sum_p H_p+\epsilon}.
\label{eq:ebpg_compact}
\end{equation}
Soft Pointing Game (SoftPG), a soft variant of the pointing game defined for this study, measures the strongest lesion-supported response relative to the global map maximum,
\begin{equation}
\mathrm{SoftPG}=\frac{\max_p(H_pM_p)}{\max_p H_p+\epsilon},
\label{eq:softpg_compact}
\end{equation}
so that a value of one means that some pixel of maximal saliency lies on an annotated lesion, and a value of 0.5 means that the strongest response on any lesion is half the global peak. SoftPG is therefore a measure of whether the map's strongest evidence is lesion-supported, not a lesion detection rate: it does not count how many lesions are found and does not penalize saliency outside the mask. Higher values are preferred for both metrics: EBPG rewards saliency energy concentrated on lesions, whereas SoftPG rewards the strongest response supported by lesions.

\paragraph{Faithfulness metrics.} Localization does not establish whether highlighted pixels materially influence the model. We therefore also use deletion and insertion analysis~\cite{petsiuk2018,gomez2022}. Pixels are ranked by saliency from highest to lowest and evaluated across 21 states from 0\% to 100\% perturbation. Let $x$ be the deterministically preprocessed input, $x_0$ a zero tensor in normalized input space, and $B_t$ the binary mask selecting the top fraction $t$ of saliency-ranked pixels. Deletion and insertion inputs are $x_{\mathrm{del}}(t)=x\odot(1-B_t)+x_0\odot B_t$ and $x_{\mathrm{ins}}(t)=x_0\odot(1-B_t)+x\odot B_t$. Because the grader outputs an unconstrained continuous severity score, responses are normalized relative to the original image and baseline. For model output $f(\cdot)$ and $\Delta=|f(x)-f(x_0)|$, the preservation score is
\begin{equation}
g(z)=\mathrm{clip}\!\left(\frac{\operatorname{sign}(f(x)-f(x_0))[f(z)-f(x_0)]}{\Delta},0,1\right).
    \label{eq:preservation_compact}
\end{equation}
Trapezoidal integration yields deletion and insertion AUC. A lower deletion AUC means that removing highly ranked pixels suppresses the response quickly, whereas a higher insertion AUC means that restoring them recovers the response quickly. We additionally report the study-specific balance $\mathrm{OA}=\mathrm{Insertion}-\mathrm{Deletion}$, but retain both constituent metrics because deletion and insertion are not exact inverses. The deletion test shares its perturbation protocol with the reliability factor $F_m$ (zero baseline, saliency-ranked removal); $F_m$ probes a single 5\% removal, whereas the deletion AUC integrates the full curve.

\paragraph{Validity and statistical testing.} Validity was tracked explicitly. A heatmap is treated as zero-valued when its maximum or total saliency does not exceed $\epsilon=10^{-8}$, and faithfulness is marked undefined for a zero map or when $|f(x)-f(x_0)|\le\epsilon$. Localization scores for a zero map are set to zero, whereas deletion, insertion, and OA are averaged only over defined curves; the number of valid curves, undefined cases, and zero-valued maps is reported for each method, so that no method appears stronger merely because difficult failures were silently excluded.

Uncertainty is reported as 95\% confidence intervals from 10{,}000 nonparametric bootstrap resamples of images. Because all methods are scored on the same 193 images, CCEM is compared on each metric with the best-performing single explainer, selected by its mean on the same cohort, through the paired per-image difference, its bootstrap interval, and a two-sided Wilcoxon signed-rank test with Holm adjustment over the five metrics.

\section{Experiments and Results}
\label{sec:results}
This section reports internal grading performance, the candidate-architecture screening, external grading on MAPLES-DR, and the quantitative and qualitative evaluation of the three baseline explainers and CCEM.

\subsection{Internal Grading Performance}
\label{sec:internal_results}
On the APTOS 2019 validation fold, the final Stage~2 model achieved a QWK of 0.93497 and an exact-grade accuracy of 0.85539 (Table~\ref{tab:validation_finalb7}). The weighted F1-score (0.84947) closely tracked accuracy, whereas the macro precision, macro recall, and macro F1-score were 0.72877, 0.68417, and 0.69936, respectively, indicating that performance was less uniform across grades, particularly for the underrepresented Mild and Severe categories.

\begin{figure*}[t]
\centering
\begin{minipage}[t]{0.30\textwidth}
    \vspace{0pt}
    \centering
    \captionof{table}{Internal validation performance of the EfficientNet-B7+CBAM ordinal regression model on the APTOS 2019 validation fold (fold 0, $n=733$). The fold was also used for checkpoint selection and threshold fitting.}
    \label{tab:validation_finalb7}
    \small
    \begin{tabular}{lc}
        \toprule
        \textbf{Metric} & \textbf{Value} \\
        \midrule
        Quadratic weighted kappa & 0.93497 \\
        Accuracy                 & 0.85539 \\
        Macro precision          & 0.72877 \\
        Macro recall             & 0.68417 \\
        Macro F1-score           & 0.69936 \\
        Weighted F1-score        & 0.84947 \\
        \midrule
        Kaggle private-test QWK  & 0.92011 \\
        \bottomrule
    \end{tabular}

    \vspace{0.35cm}

    \captionof{table}{External cross-dataset grading of the frozen Stage 2 model on all 198 MAPLES-DR images, with 95\% bootstrap confidence intervals. Referable DR: ICDR grade $\geq 2$.}
    \label{tab:external_validation}
    \small
    \setlength{\tabcolsep}{3pt}
    \begin{tabular}{lc}
        \toprule
        \textbf{Metric} & \textbf{Value [95\% CI]} \\
        \midrule
        Accuracy         & 0.470 [0.404, 0.540] \\
        QWK              & 0.559 [0.425, 0.659] \\
        Within $\pm1$ grade & 0.980 \\
        Referable sens.  & 0.828 \\
        Referable spec.  & 0.905 \\
        \bottomrule
    \end{tabular}
\end{minipage}
\hfill
\begin{minipage}[t]{0.67\textwidth}
    \vspace{0pt}
    \centering
\begingroup
\renewcommand{\sffamily}{\fontfamily{phv}\selectfont}%
\renewcommand{\bfseries}{\fontseries{b}\selectfont}%
\def\baselinestretch{1}%
\newcommand{\cpS}{\fontsize{6}{7.2}\selectfont}%
\newcommand{\cpT}{\fontsize{5}{6}\selectfont}%
\definecolor{cpPrecision}{RGB}{102,102,255}%
\definecolor{cpRecall}{RGB}{255,166,77}%
\definecolor{cpF1score}{RGB}{0,110,95}%
\definecolor{cpAx}{RGB}{75,75,75}%
\begin{tikzpicture}[x=1cm, y=1cm, font=\sffamily\cpS, text=black, line cap=butt, line join=round]
\path[use as bounding box] (0,0) rectangle (12.0900,4.9400);
\fill[cpPrecision] (0.8830,0.6200) rectangle (1.4430,4.2996);
\node[anchor=south, inner sep=0pt, font=\sffamily\cpT] at (1.1630,4.3396) {0.99};
\fill[cpRecall] (1.4830,0.6200) rectangle (2.0430,4.3098);
\node[anchor=south, inner sep=0pt, font=\sffamily\cpT] at (1.7630,4.3498) {1.00};
\fill[cpF1score] (2.0830,0.6200) rectangle (2.6430,4.3046);
\node[anchor=south, inner sep=0pt, font=\sffamily\cpT] at (2.3630,4.3446) {1.00};
\draw[cpAx, line width=0.4pt] (1.7630,0.6200) -- ++(0,-0.05);
\node[anchor=base, inner sep=0pt, font=\sffamily\cpS] at (1.7630,0.2900) {No DR (0)};
\node[anchor=base, inner sep=0pt, font=\sffamily\cpS] at (1.7630,0.0600) {\textit{n} = 361};
\fill[cpPrecision] (3.1690,0.6200) rectangle (3.7290,3.3950);
\node[anchor=south, inner sep=0pt, font=\sffamily\cpT] at (3.4490,3.4350) {0.75};
\fill[cpRecall] (3.7690,0.6200) rectangle (4.3290,2.4200);
\node[anchor=south, inner sep=0pt, font=\sffamily\cpT] at (4.0490,2.4600) {0.49};
\fill[cpF1score] (4.3690,0.6200) rectangle (4.9290,2.8036);
\node[anchor=south, inner sep=0pt, font=\sffamily\cpT] at (4.6490,2.8436) {0.59};
\draw[cpAx, line width=0.4pt] (4.0490,0.6200) -- ++(0,-0.05);
\node[anchor=base, inner sep=0pt, font=\sffamily\cpS] at (4.0490,0.2900) {Mild (1)};
\node[anchor=base, inner sep=0pt, font=\sffamily\cpS] at (4.0490,0.0600) {\textit{n} = 74};
\fill[cpPrecision] (5.4550,0.6200) rectangle (6.0150,3.3990);
\node[anchor=south, inner sep=0pt, font=\sffamily\cpT] at (5.7350,3.4390) {0.75};
\fill[cpRecall] (6.0550,0.6200) rectangle (6.6150,3.8575);
\node[anchor=south, inner sep=0pt, font=\sffamily\cpT] at (6.3350,3.8975) {0.88};
\fill[cpF1score] (6.6550,0.6200) rectangle (7.2150,3.6108);
\node[anchor=south, inner sep=0pt, font=\sffamily\cpT] at (6.9350,3.6508) {0.81};
\draw[cpAx, line width=0.4pt] (6.3350,0.6200) -- ++(0,-0.05);
\node[anchor=base, inner sep=0pt, font=\sffamily\cpS] at (6.3350,0.2900) {Moderate (2)};
\node[anchor=base, inner sep=0pt, font=\sffamily\cpS] at (6.3350,0.0600) {\textit{n} = 200};
\fill[cpPrecision] (7.7410,0.6200) rectangle (8.3010,2.0776);
\node[anchor=south, inner sep=0pt, font=\sffamily\cpT] at (8.0210,2.1176) {0.39};
\fill[cpRecall] (8.3410,0.6200) rectangle (8.9010,1.8533);
\node[anchor=south, inner sep=0pt, font=\sffamily\cpT] at (8.6210,1.8933) {0.33};
\fill[cpF1score] (8.9410,0.6200) rectangle (9.5010,1.9561);
\node[anchor=south, inner sep=0pt, font=\sffamily\cpT] at (9.2210,1.9961) {0.36};
\draw[cpAx, line width=0.4pt] (8.6210,0.6200) -- ++(0,-0.05);
\node[anchor=base, inner sep=0pt, font=\sffamily\cpS] at (8.6210,0.2900) {Severe (3)};
\node[anchor=base, inner sep=0pt, font=\sffamily\cpS] at (8.6210,0.0600) {\textit{n} = 39};
\fill[cpPrecision] (10.0270,0.6200) rectangle (10.5870,3.4112);
\node[anchor=south, inner sep=0pt, font=\sffamily\cpT] at (10.3070,3.4512) {0.75};
\fill[cpRecall] (10.6270,0.6200) rectangle (11.1870,3.3166);
\node[anchor=south, inner sep=0pt, font=\sffamily\cpT] at (10.9070,3.3566) {0.73};
\fill[cpF1score] (11.2270,0.6200) rectangle (11.7870,3.3631);
\node[anchor=south, inner sep=0pt, font=\sffamily\cpT] at (11.5070,3.4031) {0.74};
\draw[cpAx, line width=0.4pt] (10.9070,0.6200) -- ++(0,-0.05);
\node[anchor=base, inner sep=0pt, font=\sffamily\cpS] at (10.9070,0.2900) {Proliferative (4)};
\node[anchor=base, inner sep=0pt, font=\sffamily\cpS] at (10.9070,0.0600) {\textit{n} = 59};
\draw[cpAx, line width=0.4pt] (0.6200,4.3200) -- (0.6200,0.6200) -- (12.0500,0.6200);
\draw[cpAx, line width=0.4pt] (0.6200,0.6200) -- ++(-0.05,0);
\node[anchor=east, inner sep=0pt, font=\sffamily\cpS] at (0.5300,0.6200) {0};
\draw[cpAx, line width=0.4pt] (0.6200,2.4700) -- ++(-0.05,0);
\node[anchor=east, inner sep=0pt, font=\sffamily\cpS] at (0.5300,2.4700) {0.5};
\draw[cpAx, line width=0.4pt] (0.6200,4.3200) -- ++(-0.05,0);
\node[anchor=east, inner sep=0pt, font=\sffamily\cpS] at (0.5300,4.3200) {1};
\node[anchor=north, rotate=90, inner sep=0pt, font=\sffamily\cpS] at (0,2.4700) {Score (0--1)};
\node[anchor=base, inner sep=0pt, font=\sffamily\cpS] at (6.3350,4.7200) {%
\tikz[baseline=0pt]{\fill[cpPrecision] (0,-0.01) rectangle (0.35,0.16);}\hspace{0.08cm}Precision\hspace{0.45cm}%
\tikz[baseline=0pt]{\fill[cpRecall] (0,-0.01) rectangle (0.35,0.16);}\hspace{0.08cm}Recall\hspace{0.45cm}%
\tikz[baseline=0pt]{\fill[cpF1score] (0,-0.01) rectangle (0.35,0.16);}\hspace{0.08cm}F1-score%
};
\end{tikzpicture}%
\endgroup
    \captionof{figure}{Class-wise precision, recall, and F1-score of the final EfficientNet-B7 + CBAM model on the 733-image APTOS 2019 validation fold.}
    \label{fig:class_performance}

    \vspace{0.25cm}

    \resizebox{0.7\linewidth}{!}{\input{figures/confusion_tikz}}
    \captionof{figure}{Internal validation confusion matrices on the APTOS 2019 validation fold ($n=733$): (a) counts and (b) row-normalized recall.}
    \label{fig:confusion_matrix}
\end{minipage}
\end{figure*}

Figures~\ref{fig:class_performance} and~\ref{fig:confusion_matrix} locate the class-wise differences. No DR was identified correctly in 360 of 361 images (recall 99.72\%), Moderate DR in 175 of 200 (87.50\%), and Proliferative DR in 43 of 59 (72.88\%). Mild DR retained a precision of 0.7500 but a recall of only 0.4865, with 34 of 74 Mild cases assigned to Moderate DR, indicating that predictions of Mild were generally reliable although many true Mild cases were classified as another grade. Severe DR was the most challenging category, with a precision of 0.3939 and a recall of 0.3333: 16 of 39 cases were assigned to Moderate DR and 10 to Proliferative DR. Of the 106 incorrect predictions, 93 (87.74\%) differed from the reference label by only one grade, indicating that the model generally preserved the ordinal relationship among DR categories, although differentiation of neighboring grades remained difficult for the minority classes.

On the Kaggle private test set, the final prediction pipeline achieved a QWK of 0.92011 across approximately 13,000 hidden samples. Because the private-test reference labels were not released, accuracy, class-wise metrics, and a confusion matrix could not be calculated for this cohort. The private leaderboard result therefore provides the held-out estimate of ordinal grading performance, whereas the class-wise analysis above remains specific to the internal validation fold.

\subsection{Candidate Architectures}

Before the final regression pipeline was built, four candidate architectures were screened as five-class classifiers on the same validation fold (Table~\ref{tab:candidate_performance}).

\begin{table*}[t]
\centering
\caption{Candidate classification models on APTOS 2019. Validation metrics use the 733-image APTOS fold; public/private QWK values are Kaggle leaderboard results. Best in \textbf{bold}, second best \underline{underlined}.}
\label{tab:candidate_performance}
\small
\setlength{\tabcolsep}{4.0pt}
\begin{tabular}{lcccccc|cc}
\toprule
\textbf{Model} & \textbf{Acc. (\%)} & \textbf{Macro P (\%)} & \textbf{Macro R (\%)} & \textbf{Macro F1 (\%)} & \textbf{Wt. F1 (\%)} &  \textbf{Val. QWK} & \textbf{Public QWK} & \textbf{Private QWK} \\
\midrule
CheXNet & 80.90 & \underline{73.19} & 55.47 & 57.80 & 78.54 & 0.8573 & 0.6220 & 0.8451 \\
DenseNet121 & \textbf{84.72} & \textbf{76.32} & \underline{62.72} & \underline{65.62} & \underline{83.17} & \textbf{0.9030} & 0.6852 & 0.8822 \\
EfficientNet-B7 & 83.08 & 72.39 & 60.94 & 63.82 & 81.59 & 0.8907 & \underline{0.7371} & \underline{0.8824} \\
\rowcolor{Green!5} \textbf{EfficientNet-B7+CBAM (Ours)} & \underline{83.63} & 70.36 & \textbf{72.26} & \textbf{71.14} & \textbf{83.67} & \underline{0.9024} & \textbf{0.7402} & \textbf{0.8951} \\
\bottomrule
\end{tabular}
\end{table*}

DenseNet121 achieved the highest validation accuracy, macro precision, and validation QWK, whereas EfficientNet-B7+CBAM achieved the highest macro recall, macro F1-score, and weighted F1-score, and also the highest public and private leaderboard QWK. Relative to EfficientNet-B7 without CBAM, the attention-enhanced model increased macro recall from 60.94\% to 72.26\% and macro F1-score from 63.82\% to 71.14\%. Given the imbalanced grade distribution and the importance of minority-grade sensitivity, EfficientNet-B7+CBAM was selected on the basis of its validation macro recall and macro F1. This comparison served as architecture screening only; the final model was subsequently trained with the two-stage regression framework of Section~\ref{sec:training}.

\subsection{External Cross-Dataset Grading Performance on MAPLES-DR}
The frozen Stage~2 model and the APTOS-derived thresholds were applied without recalibration to all 198 MAPLES-DR images, using the deterministic view employed for the XAI analysis. As shown in Table~\ref{tab:external_validation}, the model achieved an exact-grade accuracy of 0.4697 (95\% CI 0.404--0.540) and a QWK of 0.5585 (95\% CI 0.425--0.659), 38.6 percentage points and 0.376 below the corresponding internal values. Although these differences should not be interpreted as a controlled within-dataset comparison, they indicate a marked reduction in predictive agreement when the model was transferred to the MESSIDOR-derived image distribution.

\begin{table}[t]
\centering
\caption{Confusion matrix of the frozen Stage 2 model on the 198 MAPLES-DR images (rows: MAPLES-DR reference grade; columns: predicted ICDR grade).}
\label{tab:external_confusion}
\small
\setlength{\tabcolsep}{5pt}
\begin{tabular}{lrrrrrr}
\toprule
\textbf{Reference} & \textbf{0} & \textbf{1} & \textbf{2} & \textbf{3} & \textbf{4} & \textbf{Recall} \\
\midrule
R0 (No DR), $n=14$        & \textbf{14} & 0  & 0  & 0 & 0 & 1.000 \\
R1 (Mild), $n=155$        & 75 & \textbf{64} & 16 & 0 & 0 & 0.413 \\
R2 (Moderate), $n=15$     & 0  & 3  & \textbf{12} & 0 & 0 & 0.800 \\
R3 (Severe), $n=10$       & 1  & 1  & 6  & \textbf{1} & 1 & 0.100 \\
R4A (Prolif.), $n=4$      & 0  & 0  & 2  & 0 & \textbf{2} & 0.500 \\
\bottomrule
\end{tabular}
\end{table}

Table~\ref{tab:external_confusion} locates the disagreement. The dominant error is a one-grade under-call of Mild DR: 75 of the 155 R1 images were predicted as No DR and a further 16 as Moderate, so that only 41.3\% of R1 images received the exact grade. All 14 R0 images and 12 of 15 R2 images were graded exactly, whereas the ten Severe images were distributed across all five predicted grades. Despite an exact-grade accuracy below 0.50, 97.98\% of predictions lay within one grade of the reference, which accounts for the moderate QWK, and referable DR (R2 and above) was separated from non-referable images with a sensitivity of 0.828 and a specificity of 0.905. The explanation maps generated on MAPLES-DR should therefore be understood as representations of the evidence used by the trained model under external-domain conditions: a heatmap may be faithful to the model's prediction even when that prediction disagrees with the external reference grade.

\subsection{Explainability Evaluation}
The three individual explainers exhibited distinct profiles on the 193 eligible MAPLES-DR images (Table~\ref{tab:xai_comparison}). SG-IG showed the strongest expert-mask alignment among the baselines, achieving the highest EBPG and SoftPG, consistent with fine-grained pixel-level attributions that place a larger proportion of saliency within the annotated lesions. Grad-CAM++ achieved the highest insertion AUC, but its lower EBPG and SoftPG indicate that its attribution was often distributed beyond the masks. Ada-SISE achieved the lowest deletion AUC among the baselines, but its compact activation-guided masks did not translate into stronger lesion overlap and obtained the lowest EBPG, SoftPG, insertion AUC, and OA. No individual explainer was therefore consistently superior, which motivates the adaptive fusion evaluated below.

\begin{table}[t]
\centering
\caption{Explanation quality on the 193 eligible MAPLES-DR images: mean, with the 95\% bootstrap CI beneath; EBPG and SoftPG in \%. Best in \textbf{bold}, second best \underline{underlined}. The value in parentheses after each CCEM mean is the paired per-image difference from the best single explainer in that column (the bold or underlined baseline): \textcolor{ForestGreen}{green} marks a gain, \textcolor{BrickRed}{red} a loss, and \textcolor{gray}{gray} no significant difference; $\dagger$ marks a Holm-adjusted Wilcoxon signed-rank $p<0.01$.}
\label{tab:xai_comparison}
\footnotesize
\setlength{\tabcolsep}{1.6pt}
\begin{tabular}{@{}lccccc@{}}
    \toprule
    \textbf{Method}
    & \textbf{EBPG $\uparrow$}
    & \textbf{SoftPG $\uparrow$}
    & \textbf{Del $\downarrow$}
    & \textbf{Ins $\uparrow$}
    & \textbf{OA $\uparrow$} \\
    \midrule
    Grad-CAM++
    & 5.86 & 43.19 & 0.694 & \textbf{0.704} & 0.009 \\
    & \ci{4.3, 7.6} & \ci{37.9, 48.4} & \ci{0.660, 0.728} & \ci{0.666, 0.739} & \ci{$-$0.036, 0.056} \\
    \addlinespace[2pt]
    Ada-SISE
    & 3.46 & 30.18 & \underline{0.639} & 0.646 & 0.006 \\
    & \ci{2.4, 4.7} & \ci{26.2, 34.4} & \ci{0.600, 0.678} & \ci{0.605, 0.687} & \ci{$-$0.044, 0.057} \\
    \addlinespace[2pt]
    SG-IG
    & \textbf{10.53} & \underline{56.83} & 0.644 & \underline{0.691} & \underline{0.047} \\
    & \ci{8.6, 12.6} & \ci{51.9, 61.7} & \ci{0.606, 0.681} & \ci{0.647, 0.732} & \ci{$-$0.001, 0.096} \\
    \addlinespace[2pt]
    \rowcolor{Green!5}\textbf{CCEM}
    & \underline{9.03}\dloss{$-$1.50} & \textbf{69.61}\dgain{+12.78} & \textbf{0.570}\dgain{$-$0.070} & 0.685\dns{$-$0.019} & \textbf{0.115}\dgain{+0.068} \\
    \rowcolor{Green!5}\textbf{(Ours)}
    & \ci{7.3, 10.9} & \ci{63.7, 75.3} & \ci{0.526, 0.613} & \ci{0.638, 0.730} & \ci{0.066, 0.165} \\
    \bottomrule
\end{tabular}
\end{table}

\subsubsection{Quantitative Evaluation}
\label{CCEM evaluation}
CCEM achieved the highest SoftPG of 69.61\% (95\% CI 63.7--75.3), exceeding the strongest individual baseline, SG-IG, by 12.78 percentage points on the paired comparison (95\% CI 9.2--16.5), and its SoftPG was higher than that of SG-IG on 58.5\% of images. The fused explanation therefore more consistently assigned its strongest response to at least one expert-annotated lesion region. CCEM also achieved the lowest deletion AUC of 0.570, a paired reduction of 0.070 (95\% CI 0.035--0.105) relative to Ada-SISE, the strongest baseline on this metric: removing pixels prioritized by CCEM caused the model response to decline more rapidly than removing pixels selected by any individual explainer. Its insertion AUC of 0.685 was statistically indistinguishable from that of Grad-CAM++, and the improved deletion result produced the highest OA of 0.115, exceeding SG-IG by 0.068 (95\% CI 0.025--0.111). All three gains remain significant after Holm adjustment ($p\le0.004$).

CCEM did not lead every metric. It obtained the second-highest EBPG (9.03\%), significantly below SG-IG by 1.5 percentage points but far above Grad-CAM++ and Ada-SISE. The gains are therefore in lesion-supported peak response and perturbation faithfulness, and the cost is in energy concentration, which is the expected signature of averaging maps with partially overlapping support.

The image-specific weights varied across images rather than collapsing to a constant: the mean weights were 0.30 (Grad-CAM++), 0.32 (Ada-SISE), and 0.38 (SG-IG), individual weights ranged from 0.08 to 0.77, and a single explainer received more than half of the total weight on 28\% of images. All four methods produced valid deletion, insertion, and OA curves on all 193 images, with no undefined measurements or zero-valued heatmaps, so the faithfulness means are directly comparable and are not affected by method-specific exclusion of failed cases.

\subsubsection{Qualitative Evaluation of Explanation Maps}
\begin{figure*}[t]
    \centering
    \input{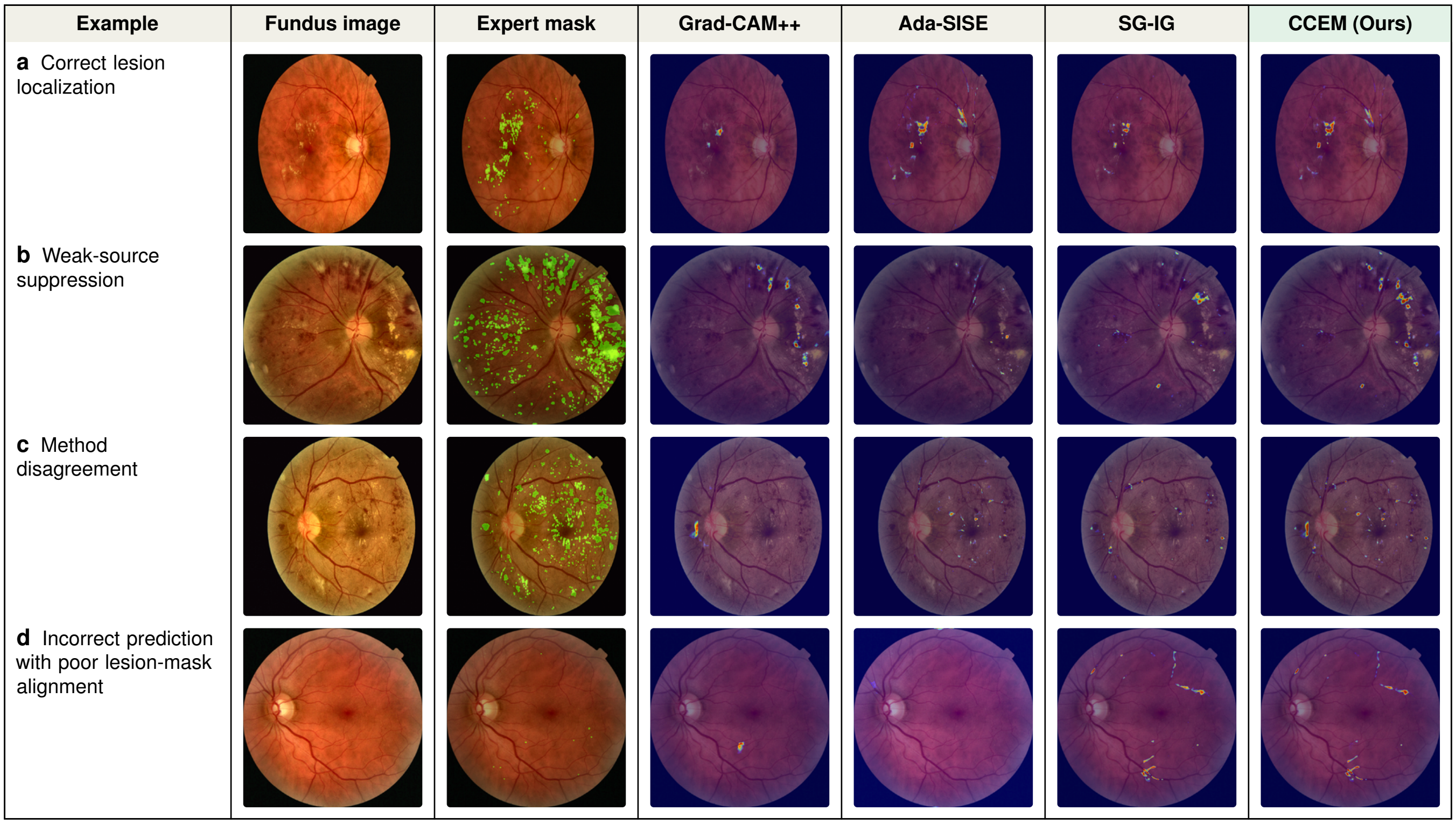}
    \caption{Qualitative comparison on representative MAPLES-DR images. Columns show the fundus image, consolidated expert lesion mask, Grad-CAM++, Ada-SISE, SG-IG, and CCEM. The examples cover agreement, disagreement, attenuation of a weak source map, and an incorrect model prediction, illustrating that explanation coherence and grading correctness are distinct.}
    \label{fig:qualitative_comparison}
\end{figure*}

Figure~\ref{fig:qualitative_comparison} presents representative MAPLES-DR examples. In the correctly localized cases, CCEM preserved lesion-associated regions supported by multiple methods while reducing isolated or weak responses. In the method-disagreement examples, Grad-CAM++ produced comparatively limited localization, whereas Ada-SISE and SG-IG highlighted broader lesion-associated evidence that was retained in the fused explanation. The prediction-error example shows that CCEM can consolidate the evidence used by the model without correcting an erroneous prediction or recovering lesion evidence absent from the source explanations; the qualitative findings should therefore be interpreted together with the quantitative results rather than as independent evidence of diagnostic correctness.

\section{Discussion}

\subsection{Grading Performance and Domain Shift}
The internal results show that the regression formulation largely preserved the ordinal structure of DR severity: most validation errors fell at adjacent grades, and the main difficulty lay in separating the two least represented grades, Mild and Severe DR, from their neighbors. Because the same APTOS fold served for checkpoint selection, threshold optimization, and reporting, the class-wise figures are internal validation results, and the hidden-test QWK is the held-out estimate.

When transferred without recalibration to MAPLES-DR, the decline was concentrated in a single systematic shift, the under-calling of Mild DR as No DR, rather than in large grade errors (Table~\ref{tab:external_confusion}). This is consistent with cross-dataset variation in image acquisition, illumination, color characteristics, and grade distribution (R1 accounts for 78\% of MAPLES-DR images but 10\% of APTOS images), and with an annotation protocol in which Mild DR may rest on a few small microaneurysms that are difficult to resolve at the grader's input resolution. Similar generalization concerns have been observed in DR systems evaluated across multiple datasets~\cite{chetoui2020,gulshan2016}. The external results should therefore be interpreted as evidence of model behavior under domain transfer rather than as an estimate of clinical diagnostic performance.

\subsection{Explanation Performance and Adaptive Fusion}
The individual explainers showed complementary behavior consistent with their attribution mechanisms: Grad-CAM++ operates on gradient-weighted convolutional activations~\cite{chattopadhay2018grad}, Ada-SISE evaluates activation-derived input masks~\cite{sudhakar2021ada}, and SG-IG combines path-integrated gradients with noise-based aggregation~\cite{smilkov2017smoothgrad,sundararajan2017axiomatic}. CCEM provided the strongest overall balance. It achieved a higher mean lesion-supported peak-response ratio than the strongest individual explainer (69.61\% versus 56.83\%), without implying a corresponding lesion detection rate, and its highest-ranked pixels were those to which the model was most sensitive. CCEM should nevertheless not be presented as uniformly superior: its contribution is the consolidation of complementary attribution evidence while reducing dependence on any single explanation mechanism. Reliability is estimated without expert lesion masks, so maps assessed as weak, diffuse, or inconsistent receive less influence on that image, while evidence supported by several reliability signals is retained.

Two features of the design bear on how the gains should be read. First, the deletion factor $F_m$ and the deletion AUC share the same perturbation protocol, so part of the deletion advantage follows from the weighting rule by construction: CCEM up-weights, on each image, the map whose top pixels the model is most sensitive to, and the deletion curve rewards exactly that. The SoftPG and OA gains are not implied by $F_m$ in the same way. Second, the present experiments evaluate the complete CCEM pipeline rather than the isolated contribution of adaptive weighting. Because equal-weight fusion and factor-wise ablations were not evaluated, the observed improvements cannot be attributed exclusively to the reliability function. The results support the complete method, but do not establish that every component is independently necessary.

\subsection{Explanation Metric Trade-offs}
The explanation metrics measure different properties and should not be reduced to a single ranking. SoftPG evaluates the strongest lesion-supported response relative to the global heatmap maximum, whereas EBPG evaluates the fraction of total saliency energy inside the expert mask; the two capture related but distinct forms of spatial correspondence~\cite{wang2020score}. A map can place a near-maximal response on a lesion while spreading most of its energy elsewhere. CCEM's broader high-saliency support, a consequence of averaging and percentile normalization, favors SoftPG for this reason. EBPG instead rewards concentrated lesion overlap, on which SG-IG's fine-grained input-level attribution performed best; CCEM ranked second, retaining much of SG-IG's lesion concentration while incorporating evidence from two activation-based explainers whose energy is less concentrated.

Deletion and insertion assess model response rather than clinical localization~\cite{petsiuk2018}. They are related but not exact inverses and may favor different saliency rankings; CCEM's leading OA was driven by its deletion improvement rather than by the highest insertion score, and OA should therefore be interpreted together with both constituent metrics. More broadly, lesion localization and model faithfulness may diverge: an explanation can faithfully represent evidence used by an incorrect model, while visual overlap with a lesion does not necessarily show that the lesion materially influenced the prediction. The prediction-error example in Figure~\ref{fig:qualitative_comparison} illustrates this distinction: fusion improves the representation of model evidence; it does not improve the underlying classifier.

\subsection{Clinical and Methodological Implications}
CCEM is intended as a model-inspection and decision-support mechanism rather than an independent diagnostic system. A reviewer can inspect the predicted grade together with the fused map and, when needed, the three source explanations and their weights, to distinguish localized retinal evidence from diffuse or anatomically implausible attribution. This presentation is most useful when the explanation is interpreted alongside the original fundus image and the model's external performance: the MAPLES-DR accuracy drop shows that a coherent heatmap does not guarantee a correct prediction, and visually plausible saliency alone is insufficient evidence of faithfulness~\cite{adebayo2018}. Full validity coverage removes one source of comparison bias, but validity is not equivalent to quality, and deletion and insertion depend on the perturbation baseline and may create out-of-distribution inputs~\cite{gomez2022,petsiuk2018}. The reported values should therefore be interpreted within the stated protocol rather than as absolute measures of clinical or causal reliability. Before deployment, the framework requires multi-center testing across cameras and populations together with ophthalmologist reader studies that measure error detection, review efficiency, and confidence calibration.

\section{Conclusion and Future Work}
\label{sec:conclusion}
This study proposed CCEM, an image-specific reliability-weighted fusion of complementary post-hoc explanations, and evaluated it on a two-stage ordinal DR grader against expert lesion annotations. On the 193-image external MAPLES-DR cohort, CCEM achieved the best SoftPG, deletion AUC, and deletion--insertion balance, each significantly better than the strongest single explainer, while SG-IG retained the best energy-based localization (EBPG). The grader reached a QWK of 0.920 on the hidden Kaggle test set and strong internal APTOS validation performance, but its reduced MAPLES-DR accuracy and QWK showed incomplete cross-domain generalization. The result supports multi-explainer fusion as a useful inspection strategy without implying uniform superiority over every source method.

The study remains limited by one finalized training split, a relatively small external cohort, a consolidated rather than lesion-specific expert mask, and the absence of equal-weight or factor-wise CCEM ablations. Future work should repeat training across seeds and architectures, evaluate fixed/equal-weight fusion and individual reliability factors, extend validation to larger multi-center retinal datasets and Vietnamese clinical data, and test a clinician-facing interface in which users can inspect, question, reject, or correct explanation evidence.


\balance
\bibliographystyle{ACM-Reference-Format}
\bibliography{references}

\end{document}